\documentclass[a4paper]{llncs}

\usepackage[american,english]{babel}
\usepackage[hidelinks]{hyperref}
\usepackage{amsmath}
\usepackage{amssymb}
\usepackage[misc]{ifsym}
\usepackage{newtxtext}
\usepackage{newtxmath}
\usepackage{wtref}
\usepackage{enumitem}
\usepackage{graphicx}
\usepackage{tikz}
\usepackage{tcolorbox}
\usepackage{acro}
\usepackage{pgfplots}
\usepackage{pgfplotstable}
\usepackage[labelfont=bf,font=small,skip=4pt,labelsep=period]{caption}
\usepackage{docmute}

\title{%
  A Quantitative Evaluation of Natural Language Question
  Interpretation for Question Answering Systems}

\author{%
  Takuto~Asakura\inst{1}\corrAuthor
  \and
  Jin-Dong~Kim\inst{3} \and
  Yasunori~Yamamoto\inst{3} \and
  Yuka~Tateisi\inst{4} \and
  Toshihisa~Takagi\inst{2}}

\institute{%
  Department of Informatics, SOKENDAI, Tokyo, Japan \\
  \email{asakura@nii.ac.jp}
  \and
  Department of Bioinformatics and Systems Biology,
  The University of Tokyo, Tokyo, Japan \\
  \email{tt@bs.s.u-tokyo.ac.jp}
  \and
  Database Center for Life Science, Chiba, Japan \\
  \email{\{jdkim,yy\}@dbcls.rois.ac.jp}
  \and
  National Bioscience Database Center, Tokyo, Japan \\
  \email{tateisi@biosciencedbc.jp}}

\input{general.sty}
\input{acros.sty}

\begin{document}

\maketitle

\begin{abstract}
Systematic benchmark evaluation plays an important role in the process of
improving technologies for \ac{qa} systems. While currently there are a number
of existing evaluation methods for \ac{nl} \ac{qa} systems, most of them
consider only the final answers, limiting their utility within a black box
style evaluation. Herein, we propose a subdivided evaluation approach to enable
finer-grained evaluation of \ac{qa} systems, and present an evaluation tool
which targets the \ac{nlq} interpretation step, an initial step of a \ac{qa}
pipeline. The results of experiments using two public benchmark datasets
suggest that we can get a deeper insight about the performance of a \ac{qa}
system using the proposed approach, which should provide a better guidance for
improving the systems, than using black box style approaches.
\end{abstract}


\acresetall

\section{Introduction}

Recently, \ac{ld} has been recognized as an emerging standard for the
integration of databases and the number of \acp{rkb} is rapidly
increasing~\cite{bizer2009,lehmann2015,vrandevcic2012}. While \ac{sparql} is
recognized as a standard tool for treating \acp{rkb}, authoring queries in
\ac{sparql} is not so easy especially for non-technicians~\cite{harris2013}.
For this reason, systems that allow users to search \acp{rkb} through
\acp{nlq}, the so-called \ac{qa} systems, are recognized as being highly
useful. In particular, \ac{qa} systems generating \ac{sparql} queries from
\acp{nlq} are called \ac{sqa} systems (the exact definition is available in a
study by H\"offner~\cite{hoffner2017}).

Benchmarking evaluations play an important role in improving \ac{sqa} systems.
While there are a number of existing evaluation methods, these methods
essentially evaluate only the final answers per input \acp{nlq}~%
\cite{lopez2013}. However, since \ac{sqa} systems have to involve various
processes (e.g.,~parsing \acp{nlq} and finding \acp{uri} of entities),
evaluations with only the final answers will not highlight the reasons why the
results of the evaluations are unexpected. This limitation is considerably
inconvenient for developers who are trying to improve their systems.

With this observation in mind, we propose a new evaluation method for \ac{sqa}
systems with which we aim to provide subdivided evaluation results rather than
checking only the final answers. One of the possible evaluation directions is
to focus on how valid logical expressions can be generated from the input
\acp{nlq}. In other words, evaluations on the \ac{nlq} interpreter, a module of
\ac{sqa} systems, are useful. As a similar attempt, Abacha et al.\@ analyzed
the false cases of their \ac{sqa} system and classified them into (1)~errors
associated with the answer type and (2)~errors associated with relation
extraction~\cite{abacha2012}. Herein, we employ this approach to evaluate
\ac{nlq} interpreters and present a calculation scheme for quantitative
evaluation. The method has been implemented for evaluating a module of
\acs*{okbqa}, a highly generalized \ac{sqa} framework (see Section
\secref{okbqa}), and it is available at
\url{https://github.com/wtsnjp/eval_tgm}. This program will be the first module
of the subdivided evaluation framework for the entire \ac{sqa} system.

\section{Benchmark Datasets and \acs*{sqa} Systems}

\subsection{Datasets}
\seclabel{datasets}

There are several famous benchmarks for \ac{qa} systems, e.g.,
\textsc{WebQuestions}~\cite{berant2013}, \textsc{SimpleQuestions}~%
\cite{bordes2017} and \textsc{BioASQ}~\cite{balikas2015}. Although such
datasets contain thousands of question--answer pairs, which are also annotated
with some other information, these are not suitable for our purpose because
nothing that expresses the logical structures of the questions is contained in
these datasets.

One of the formal languages or logical expressions is $\lambda$-Calculi, and
\textsc{Free917}~\cite{cai2013} has 917 pairs of \acp{nlq} and corresponding
$\lambda$-Calculuses. However, using \ac{sparql} as a logical expression is
much more reasonable for our tasks. This is because questions that can be
annotated with \ac{sparql} clearly lie in the scope of the \ac{sqa} systems,
and both \ac{sparql} queries or some similar expressions collected from
datasets and those generated by \ac{sqa} systems can be treated in exactly the
same way (e.g.,~both \ac{sparql} queries out of databases and generated
\ac{sparql} queries can be parsed by the same parser). For these reasons, two
datasets comprising pairs of \acp{nlq} and \ac{sparql} queries are chosen to be
the benchmark datasets for our evaluation.

\subsubsection{\acs*{qald}}
\seclabel{qald}

As one of the most well-known evaluation tasks,
\ac{qald}\footnote{\url{https://qald.sebastianwalter.org/}}~\cite{lopez2013,%
cimiano2013,unger2014,unger2015,unger2016,usbeck2017} contains a number of
questions annotated with equivalent \ac{sparql}
queries~(Table~\tabref{qald-overview}). Some of the datasets (e.g.,~those
named \texttt{multilingual}) contain not only English questions but also
questions in several other languages; however, we used only the English
questions. The questions in the datasets that do not contain \ac{sparql}
queries (i.e.,~\texttt{hybrid} datasets from \ac{qald}-4--7) are annotated
with pseudo queries instead. These are quite similar to \ac{sparql} queries but
can contain free text as the node of the triples, which makes the triple
patterns of the pseudo queries different from those of the actual queries.
Therefore, these datasets are inappropriate for our evaluation.

Practically, \texttt{datacube} from \ac{qald}-6 and \texttt{largescale-test}
from \ac{qald}-7 are also inappropriate for our purpose. The \ac{sparql}
queries in the former datasets comprise a lot of extended syntaxes; thus, it is
difficult for us to treat them as valid \ac{sparql} queries. The latter dataset
has 2 million questions, but this dataset is mechanically generated by an
algorithm using the questions available in the training dataset. Thus,
\texttt{largescale-test} contains a large number of similar questions; hence,
we have chosen to skip the dataset in our evaluation.

Moreover, although each newer dataset is not a proper superset of the dataset
for the previous tasks, many questions appear multiple times throughout the
datasets. Using the same questions more than once can cause bias in the
evaluation results; hence, such occurrences should be avoided.

Due to these reasons, quite a few questions were discarded from our
evaluations, but we could still obtain a reasonable number of questions
annotated with appropriate \ac{sparql} queries. The exact number of questions
used for our experiments and some of the basic analyses conducted on them will
be presented in Section~\secref{results}.
\begin{table}[bt]
\let\ok\checkmark
\centering\small
\caption{Overview of datasets provided by \acs*{qald}. The rightmost column
shows whether each dataset was used for our experiments (see Section~%
\secref{qald} for detailed reasons).}
\tablabel{qald-overview}
\begin{tabular}{llr|cc|c}
Challenge & Dataset & Size & Question (en) & \acs*{sparql} query & Used \\
\hline
QALD-1 & \texttt{dbpedia-\{train,test\}}      & 100 & \ok & \ok & \ok \\
       & \texttt{musicbrainz-\{train,test\}}  & 100 & \ok & \ok & \ok \\
\hline
QALD-2 & \texttt{dbpedia-\{train,test\}}      & 200 & \ok & \ok & \ok \\
       & \texttt{musicbrainz-\{train,test\}}  & 200 & \ok & \ok & \ok \\
       & \texttt{participants-challenge}      & 7   & \ok & \ok & \ok \\
\hline
QALD-3 & \texttt{esdbpedia-\{train,test\}}    & 100 & \ok & \ok & \ok \\
       & \texttt{dbpedia-\{train,test\}}      & 199 & \ok & \ok & \ok \\
       & \texttt{musicbrainz-\{train,test\}}  & 199 & \ok & \ok & \ok \\
\hline
QALD-4 & \texttt{multilingual-\{train,test\}} & 250 & \ok & \ok & \ok \\
       & \texttt{biomedical-\{train,test\}}   & 50  & \ok & \ok & \ok \\
       & \texttt{hybrid-\{train,test\}}       & 35  & \ok &     &     \\
\hline
QALD-5 & \texttt{multilingual-\{train,test\}} & 350 & \ok & \ok & \ok \\
       & \texttt{hybrid-\{train,test\}}       & 50  & \ok &     &     \\
\hline
QALD-6 & \texttt{multilingual-\{train,test\}} & 450 & \ok & \ok & \ok \\
       & \texttt{hybrid-\{train,test\}}       & 75  & \ok &     &     \\
       & \texttt{datacube-\{train,test\}}     & 150 & \ok & \ok &     \\
\hline
QALD-7 & \texttt{multilingual-\{train,test\}} & 314 & \ok & \ok & \ok \\
       & \texttt{hybrid-\{train,test\}}       & 150 & \ok &     &     \\
       & \texttt{largescale-train}            & 100 & \ok & \ok & \ok \\
       & \texttt{largescale-test}             & 2M  & \ok & \ok &     \\
       & \texttt{en-wikidata-\{train,test\}}  & 150 & \ok & \ok & \ok \\
\end{tabular}
\end{table}

\subsubsection{\acs*{lc-quad}}

\ac{lc-quad}\footnote{\url{https://figshare.com/projects/LC-QuAD/21812}}%
~\cite{trivedi2017} is a newer dataset that also contains 5,000 pairs of
questions in English and \ac{sparql} queries. This dataset is for machine
learning-based \ac{qa} approaches. It is also useful for our evaluation owing
to its size and complexity. Unlike the \texttt{largescale-test} dataset from
\ac{qald}-7, \ac{lc-quad} was carefully created to exclude questions that are
similar to each other.

\subsection{The \acs*{sqa} Framework: \acs*{okbqa}}
\seclabel{okbqa}

The \ac{okbqa}\footnote{\url{http://www.okbqa.org/}} community has been
developing the \ac{okbqa} framework by modularizing general \ac{sqa} systems so
that each module can be developed independently by experts in each of the
related technologies~\cite{kim2016,kim2017}. Hence the framework share the
same goal with our evaluation method. The main part of the framework or the
\ac{sqa} workflow comprises the following modules~(Figure~%
\figref{okbqa-workflow}).
\begin{figure}
\small\centering
\begingroup\input{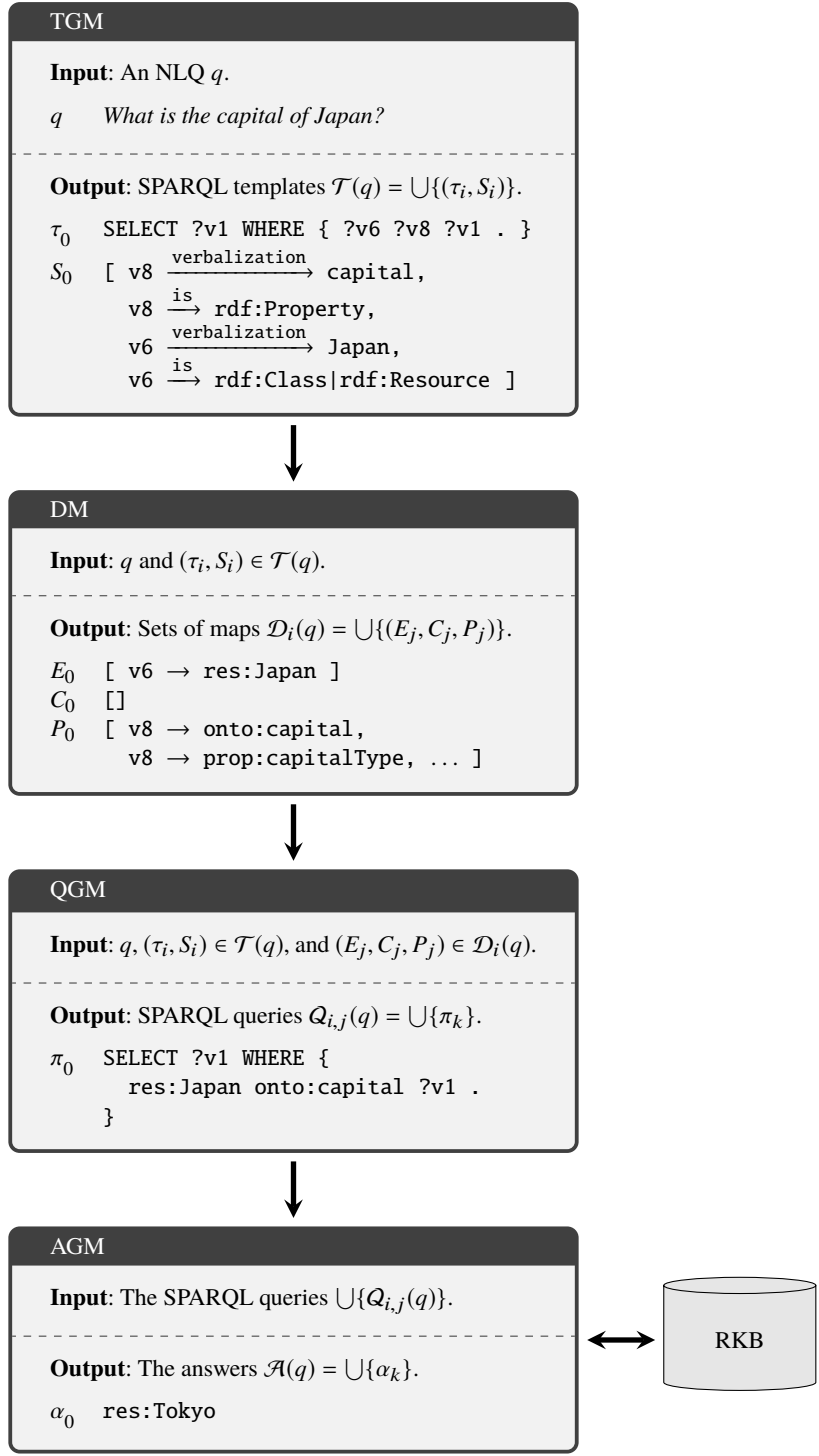}\endgroup
\caption{The workflow of the \ac{okbqa} framework with the example outputs for
the question ``What is the capital of Japan?'' The prefixes used in this figure
are summarized in Table \tabref{prefixes}. Only the essences of the inputs or
outputs of each module are shown here; hence, the actual \acp{api} allow for
attaching additional information to the inputs. For instance, many components
of the outputs are annotated with a score so that \acsp*{agm} will be able to
select or filter the candidate \ac{sparql} queries.}
\figlabel{okbqa-workflow}
\end{figure}
\begin{itemize}

\item \textbf{\acp{tgm}} take an \ac{nlq} ($q$) as their input and return a
list of \ac{sparql} templates $\mathcal{T}(q)$, which are pairs of template
queries and sets of slots $(\tau_i, S_i)$. Here, a template query, $\tau_i$, is
similar to a \ac{sparql} query, but all components of its triples are unbounded
variables, and the set of slots $S_i$ holds the descriptions of the variables.
Generally, \ac{sparql} templates represent the semantic structures of the
questions~\cite{unger2012}. Therefore, typical \acp{tgm} create them by using
some \ac{nlp} techniques.

\item \textbf{\acp{dm}} receive a \ac{sparql} template
$(\tau_i,S_i)\in\mathcal{T}(q)$ and identify resources corresponding to each of
the slots in $S_i$. More specifically, a result of \acp{dm} $\mathcal{D}_i$ is
a set of three tuples $(E_j, C_j, P_j)$, where each tuple is a list of slots to
\acp{uri} mappings for entities, classes, and properties, respectively.
Normally, \acp{dm} require \ac{rkb}-dependent information in addition to their
input from \acp{tgm}.

\item \textbf{\acp{qgm}} generate actual \ac{sparql} queries
$\mathcal{Q}_{i,j}(q)$ based on a template ($(\tau_i, S_i) \in \mathcal{T}(q)$)
and three tuples of mappings $(E_j, C_j, P_j) \in \mathcal{D}_i(q)$. This
module tends to generate many \ac{sparql} queries for each input template.

\item \textbf{\acp{agm}} query all specified \acp{rkb} using \ac{sparql}
queries generated by a \ac{qgm} and return the list of final answers
$\mathcal{A}(q)$ for the question $q$. The role of this module is not only
collecting results from \acp{rkb} but also selecting and filtering the input
\ac{sparql} queries.

\end{itemize}

Because of this modular architecture, the users of the \ac{sqa} system can
freely choose the exact implementation to execute as each module in the
\ac{sqa} workflow, and for easing the collaboration, every module
implementation has \ac{rest} services to exchange their inputs/outputs. It is
worth noting that the framework is particularly useful for our subdivided
evaluations because it is helpful to clarify that the scopes of the evaluations
and the evaluators developed for each module can be easily applied to multiple
implementations.

Now, we can clearly declare the objective of this study, which is to define and
develop an evaluation for the \acp{tgm} of the \ac{okbqa} framework.
Currently, there are two \ac{tgm} implementations for English \ac{qa}:
Rocknrole and \acs*{lodqa}. We evaluated both using our evaluation method.

\subsubsection{Rocknrole}

Rocknrole\footnote{\url{http://repository.okbqa.org/components/21}}~\cite{unger2012}
is a rule-based \ac{tgm} implementation. The approach of this implementation is
quite simple: first, the input question is parsed by the general \ac{nl} parser
included in Stanford CoreNLP~\cite{manning2014} and then converted to a
\ac{sparql} template query using predefined rules (e.g.,~the node \textit{who}
is renamed to \texttt{AGENT}). Because of its \ac{sparql} templates generation
scheme, the quality and coverage of the output are dependent on the rules.

\subsubsection{\acs*{lodqa}}

\ac{lodqa}\footnote{\url{http://lodqa.org/}}~\cite{kim2013,cohen2013} is one
of the \ac{sqa} systems that generate \ac{sparql} queries. \ac{lodqa} has a
modular architecture that resembles the \ac{okbqa} framework, and one of the
modules of the system called \textit{Graphicator} can be used as a \ac{tgm}
implementation solely by adjusting the output to conform to the \ac{tgm}
specification. The backend for deep relation extraction of the system is
Enju~\cite{miyao2008}, a state-of-the-art \ac{hpsg}-based English parser.
Then, a graph conversion algorithm involving tasks such as entity recognition
and graph simplification is executed.

\section{Methods}

Figure~\figref{evaluator} shows the overview of the calculation scheme in our
evaluator. The detailed information about each process is described in this
section.
\begin{figure}
\small\centering
\begingroup\input{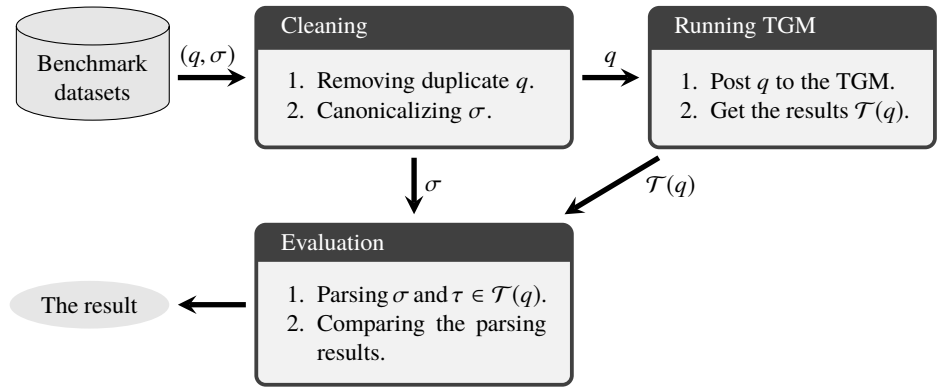}\endgroup
\caption{The calculation scheme of the evaluation. The inputs of our evaluator
are pairs of the \acp{nlq} and corresponding \ac{sparql} queries $(q, \sigma)$.
After executing all the processes, the evaluator outputs the evaluation results
for the \ac{tgm}.}
\figlabel{evaluator}
\end{figure}

\subsection{Preparation}

Before using the pairs of the \acp{nlq} and \ac{sparql} queries from the
datasets of \ac{qald} and \ac{lc-quad}, two simple processes were required to
be applied to these queries for cleaning purposes. First, due to the existence
of duplicate questions in both datasets, it was necessary to remove those
duplicate questions from the input of the evaluator. Duplicate \acp{nlq} that
were paired with different \ac{sparql} queries existed in the \ac{qald}
datasets, and thus, the queries from the newer datasets may be more effective
compared with those contained in the older ones. Consequently, the pairs from
the newer datasets were selected in such cases. Second, the \ac{sparql} queries
from both datasets included extended syntaxes of
Virtuoso,\footnote{\url{https://virtuoso.openlinksw.com/}} so many of these
queries did not satisfy the specification of \ac{sparql} 1.1,\footnote{%
\url{https://www.w3.org/TR/sparql11-query/}} which may cause parse failure of
the queries. To avoid such trivial errors, these invalid queries were modified
to conform to the official specification before the evaluation. This process
was performed in an ad-hoc manner: we made the modification using regular
expressions in our formatter scripts to create the input data
files~(Figure~\figref{regulation}).
\begin{figure}[tb]
\centering\small
\begingroup\input{regulation.def}\endgroup
\caption{Examples of the substitutions to modify the \ac{sparql} queries
containing extended syntaxes. Here, the names of variables added by our scripts
have prefix \texttt{tgm\_eval\_} to avoid any name confliction.}
\figlabel{regulation}
\end{figure}

After filtering and modification, the remaining \acp{nlq} constituted our
benchmark datasets. These datasets were then fed to the two \acp{tgm} and the
outputs were retrieved through their \ac{rest} services~(Table~\tabref{rest}).
As a result, we obtained the pairs of \acp{nlq} and corresponding \ac{sparql}
queries $(q, \sigma)$ from the datasets as well as the pairs of template
queries and sets of slots $\mathcal{T}(q) = (\tau, S)$ from the two \acp{tgm}.
As mentioned in Section \secref{okbqa}, the specification of the \ac{okbqa}
framework allows \acp{tgm} to output multiple \ac{sparql} templates for an
\ac{nlq}, but both Rocknrole and \ac{lodqa} currently output at most one
template. Hence, we did not consider the cases of multiple \ac{tgm} outputs
from an input for our experiments.
\begin{table}[tb]
\centering\small
\caption{The \ac{rest} services of the \acp{tgm}. When users send POST requests
on \ac{http}, the services will run the \ac{tgm} internally and return the
results to users. While Rocknrole supports English and Korean, \ac{lodqa}
currently supports only English.}
\tablabel{rest}
\begin{tabular}{lll}
\acs*{tgm} name & Service \ac{url} & Languages \\
\hline
Rocknrole & \url{http://ws.okbqa.org:1515/templategeneration/rocknrole}
  & en, ko \\
\acs*{lodqa} & \url{http://lodqa.org/template.json} & en \\
\end{tabular}
\end{table}

As the last step for our preparation, we input both the \ac{sparql} queries
$\sigma$ from the datasets and template queries $\tau$ from the \acp{tgm}
outputs into a \ac{sparql} parser. Specifically, we used an internal
\ac{sparql} parser from RDFLib\footnote{%
\url{https://rdflib.readthedocs.io/en/stable/}} solely because it is convenient
and fast to call this parser from our evaluation scripts written in Python. The
parser outputs the internal expressions of the parsed \ac{sparql} queries,
similar to \ac{sparql} syntax expressions or \ac{sparql} Algebra, and we can
easily extract the logical structure of the \ac{sparql} queries, e.g., the
triple patterns and the length of the answers from such queries. Technically,
the parser must be initialized with a few namespace mappings~(Table~%
\tabref{prefixes}) because some of the \ac{sparql} queries in the datasets do
not have an explicit declaration of the prefixes.
\begin{table}[t]
\centering\small
\caption{The namespace mappings used to initialized the internal \ac{sparql}
parser of RDFLib in our experiments. Since our evaluation for \acp{tgm} does
not check any particular resource, the exact \acp{uri} shown here is not so 
important.}
\tablabel{prefixes}
\begin{tabular}{ll}
Prefix & Partial \ac{uri} string \\
\hline
\texttt{dc}   & \url{http://purl.org/dc/elements/1.1/}            \\
\texttt{foaf} & \url{http://xmlns.com/foaf/0.1/}                  \\
\texttt{obo}  & \url{http://purl.obolibrary.org/obo/}             \\
\texttt{onto} & \url{http://dbpedia.org/ontology/}                \\
\texttt{owl}  & \url{http://www.w3.org/2002/07/owl#}              \\
\texttt{prop} & \url{http://dbpedia.org/property/}                \\
\texttt{rdf}  & \url{http://www.w3.org/1999/02/22-rdf-syntax-ns#} \\
\texttt{reds} & \url{http://www.w3.org/2000/01/rdf-schema#}       \\
\texttt{res}  & \url{http://dbpedia.org/resource/}                \\
\texttt{xsd}  & \url{http://www.w3.org/2001/XMLSchema#}           \\
\end{tabular}
\end{table}

\subsection{Evaluation}

The goal of our evaluation is to judge the qualities of the outputs of
\acp{tgm} independently from the other part of the \ac{sqa} systems (namely,
\acp{dm}, \acp{qgm}, and \acp{agm}). For this reason, we leave the analyses on
the sets of slots $S$ and comparing graph similarity to another step, which
will follow the \ac{tgm} evaluation (see Section~\secref{unsuitable}). Thus,
our method focus on foundational analyses particularly on the template queries
$\tau$.

To achieve our goal for subdivided evaluation, we established six evaluation
criteria based on three aspects: (1)~robustness of a \ac{tgm}, (2)~validity of
query types and the ranges (i.e.,~lengths and offsets) expressed in template
queries, and (3)~accuracy of the graph patterns in template queries. For each
aspect, two actual evaluation criteria have been developed, as listed in
Table~\tabref{criteria}. Our evaluator is implemented to check every output of
a \ac{tgm} via a comparison with the corresponding queries from the benchmark
dataset to verify whether any of the six criteria are met. If an output clears
all the criteria, it is determined to be \emph{good}.
\begin{table}
\centering\small
\let\ok\checkmark
\caption{Overview of the evaluation criteria. Our evaluator checks every output
of a \ac{tgm} to evaluate whether the output has any problem when compared with
each criterion in the exact order shown here. If errors are found, the error
that is found first is considered. For details of each criteria, see Section~%
\secref{robustness}--\secref{gp}.}
\tablabel{criteria}
\begin{tabular}{rlll}
No.\@ & Evaluation criteria & Aspects & Level \\
\hline
1 & \ac{tgm} failure    & Robustness             & Critical \\
2 & Syntax              & Robustness             & Critical \\
3 & Question type       & Query types and ranges & Critical \\
4 & Disconnected target & Graph patterns         & Critical \\
5 & Wrong range         & Query types and ranges & Notice   \\
6 & Disconnected triple & Graph patterns         & Notice   \\
\end{tabular}
\end{table}

For the convenience of the developers of \acp{tgm}, we also categorized the
criteria into two severity levels: \emph{critical} and \emph{notice}. The
difference between the two levels is related to the impact on the general
evaluation criteria, e.g., recall and precision, which are widely used for
evaluation in information systems, e.g., \ac{qald} campaign~\cite{lopez2013}:
\begin{align*}
\func{Recall}(q)
&= \frac{\text{Number of correct system answers for $q$}}
        {\text{Number of gold standard answers for $q$}}, \\
\func{Precision}(q)
&= \frac{\text{Number of correct system answers for $q$}}
        {\text{Number of system answers for $q$}}.
\end{align*}
For instance, if a \ac{sparql} template, $\mathcal{T}(q)$, is judged to have a
critical error, it means there is no chance of a correct answer, regardless of
the performance of other modules. Additionally, the contribution of the
template for precision and recall will be zero. On the contrary, if a
$\mathcal{T}(q)$ is determined to have a notice problem, it means there is
still a chance for obtaining correct answers, irrespective of how low this
chance is. Thus, its contribution to precision and recall may not be zero.

\subsubsection{Robustness}
\seclabel{robustness}

The first two steps for our evaluation are related to the robustness of the
\acp{tgm}. We call it a \emph{\ac{tgm} failure} error if the status code of the
\ac{http} response from the \ac{rest} service is not 200, which means that
somehow the \ac{tgm} did not return normal results (e.g.,~a kind of internal
error was raised for the input). If the \ac{rest} service would have returned a
\ac{sparql} template, $\mathcal{T}(q)$, the parsing result of the template
query $\tau$ would have been checked for the next step. As explained in
Section~\secref{okbqa}, a valid template query is also valid as a \ac{sparql}
query. Therefore, a template query for which the parsing result is ``syntax
error'' has \emph{syntactic problems}. Since both the problems concerning the
criteria explained here will make it difficult to follow the steps of the
framework, these problems are classified into \emph{critical} errors.

\subsubsection{Query Types and Ranges}

Generally, \acp{nlq}, which can be treated as the inputs of \ac{sqa} systems,
are roughly categorized into the following question types~\cite{hoffner2017,%
balikas2015}.
\begin{itemize}

\item \textbf{Yes/no questions} are questions that can be answered simply as
``\emph{yes}'' or ``\emph{no}'' (e.g.,~``Are there drugs that target the
Protein kinase C$\beta$ type?''). These questions can be converted directly to
\ac{sparql} queries using the \texttt{ASK} form, i.e., the so-called ask
queries.

\item \textbf{Factoid questions} require one or more entities as their answers
  (e.g.,~``Which drugs have no side-effects?''). The aim of these questions can be
easily reached by the most common \ac{sparql} queries using the \texttt{SELECT}
form, namely select queries. Sometimes, the questions that require more than
one answer are distinguished from this category (often referred to as ``list
questions''), but we did not separate that category from factoid questions
because it is a trivial matter for \ac{sparql} queries.

\item \textbf{Summary questions} are questions that are not categorized into
any of the previous types (e.g.,~``Why do people fall in love?''). The
questions typically require text as the answers; therefore, the questions
belonging to this class are out of the scope of \ac{sparql} queries.

\end{itemize}

In summary, an input \ac{nlq} of an \ac{sqa} system is basically classified
into yes/no questions or factoid questions, which can be easily detected by
checking the type of the annotated \ac{sparql} query. Using this idea, we
determined whether \acp{tgm} can accurately recognize the question types by
comparing the parsing results of the dataset \ac{sparql} queries $\sigma$ and
the template queries $\tau$: in the case wherein one of the queries is an ask
query and the other is not and vice versa, we judged that the \ac{tgm} failed
to recognize the \emph{question type} of the \ac{nlq}. This error is
\emph{critical} because incorrect types of queries always return the wrong type
of answers.

Focusing on the factoid questions, it is worth considering more detailed
classification among them. There are some questions, e.g., ``Who are the four
youngest MVP basketball players?'' wherein the number of answers have important
meanings. Moreover, the positions or the offsets of the answers (i.e., the
positions in the sorted lists of answer candidates) are important in some
questions, e.g., a \ac{sparql} query corresponding to a question such as ``What
is the largest country in the world?'' should consider the first entity from
the (sorted) candidate entities while it is desirable for a \ac{sparql} query
to consider the second question ``What is the second highest mountain on
Earth?'' Herein, we refer to these questions as \textbf{range-specified factoid
questions}. The ranges of the answers, a pair of length $l$ and starting
position $s$, can be expressed in a \ac{sparql} query by adding clauses, such
as ``\texttt{LIMIT $l$ OFFSET $(s-1)$}.'' Thus, we checked every template query
$\tau$ that correctly recognized the original question as a factoid question
(if and if only one of the queries $\sigma$ and $\tau$ was not an ask query)
and had the appropriate range specification in the query again using the
parsing result of both the $\sigma$ and $\tau$ queries. If a range $(l, s)$
explicitly appeared in the \ac{sparql} query $\sigma$ and either one of the
lengths and starting positions in the template query $\tau$ were different from
$l$ and $s$, respectively, the template was judged to have \emph{wrong range}
for the answers. Since adding the range annotations to template queries is an
optional behavior of \ac{tgm} to increase the precision, this criterion is
rightfully categorized to \emph{notice}.

\subsubsection{Graph Patterns}
\seclabel{gp}

The basic structures of \ac{sparql} queries (select queries in a precise sense)
can briefly be expressed as follows~\cite{le2012}.
\begin{quote}
\texttt{SELECT \Meta{result description} WHERE \Meta{graph patterns}},
\end{quote}
where the part \Meta{graph patterns} is a set of triple patterns and
\Meta{result description} is an enumeration of the variables requested to be
solved by the queried \ac{rdf} store, possibly with some arithmetic operators.
Herein, we simply call these variables ``target variables.''

For each template query that has the form of select queries, we checked whether
all target variables appeared in the \Meta{graph patterns}. If there were
target variables that did not exist in the \Meta{graph patterns}, then an alert
was generated as a \emph{disconnected target} error~(Figure~
\figref{graph-examples}A, \figref{graph-examples}B). This is one of the
\emph{critical} errors because the queries having this problem will retrieve
nothing for those targets.
\begin{figure}[tb]
\centering\small
\begingroup\input{graph-examples.def}\endgroup
\caption{Examples of template queries. A.~The target variable \texttt{?v4} is a
disconnected target because it does not appear in the graph patterns. B.~The
target variable \texttt{?v1\_count} does not appear in the graph patterns, but
the variable is bound as the number of \texttt{?v1} and \texttt{?v1} appears in
the patterns ``\texttt{?v1 ?v2 ?v3}.'' Therefore, this template query does not
have any problem. C.~This query has a disconnected triple: while the first two
triples in the graph patterns have a connection to the target variable
\texttt{?v1}, the last triple does not.}
\figlabel{graph-examples}
\end{figure}

The last criterion of our evaluation is related to another kind of analysis on
\Meta{graph patterns}. If there are triple patterns that are disconnected from
any target, this can be a cause of reducing the meaningful results from the
final answers for nothing. Thus, our evaluator found template queries having
those triples that were highly unnecessary, which were considered as
\emph{disconnected triple notifications}~(Figure~\figref{graph-examples}C).

\section{Results}
\seclabel{results}

After removing the duplicate \acp{nlq} from the datasets, we obtained 1,011
pairs of \acp{nlq} and \ac{sparql} queries from the \ac{qald} datasets and
4,977 pairs from \ac{lc-quad}~(Figure~\figref{questions}). For the datasets
pertaining to \ac{qald} and \ac{lc-quad}, the ratio of yes/no questions was
8.4\% and 7.4\%, respectively. In contrast, 7.0\% of the factoid questions from
\ac{qald} are range-specified, which are useful to check the existence of the
\emph{wrong range} criterion, but there are no range-specified factoid
questions in \ac{lc-quad}. All pairs were entered into our evaluator, and
every \ac{nlq} in them was successfully sent to the \ac{rest} services of the
\acp{tgm}. Likewise, every query in the pairs was parsed by the \ac{sparql}
parser in RDFLib without any issues owing to the normalization adopted in our
formatter.
\begin{figure}[p]
\centering\small
\begingroup\input{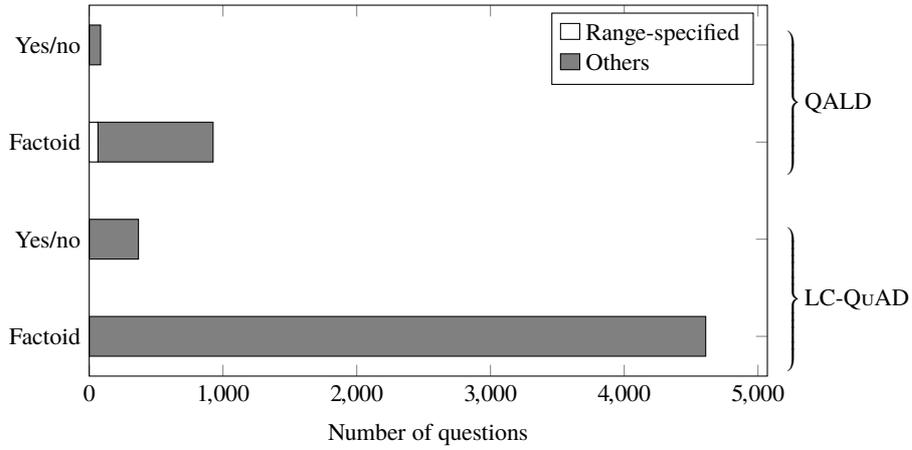}\endgroup
\caption{Sizes of the datasets used in our experiments. The datasets from
\ac{qald} contain 85 yes/no questions and 926 factoid questions, 65 of which
were range-specified. On the contrary, the dataset from \ac{lc-quad} comprises
368 yes/no questions and 4,609 factoid questions, none of which had a range
specification.}
\figlabel{questions}
\end{figure}

Table~\tabref{errors} summarizes the problems of \ac{tgm} outputs we found
through evaluation using the whole datasets (5,988 questions in total). These
problems are classified into the six criteria, as defined earlier. Overall,
9.9\% of the \ac{sparql} templates produced by Rocknrole had \emph{critical}
errors and 48.7\% of these were alerted to have \emph{notice} problems.
Similarly, 7.8\% of the templates generated by \ac{lodqa} had \emph{critical}
errors and 1.1\% of these were alerted to have \emph{notice}
problems~(Figure~\figref{errors}).
\begin{table}[p]
\centering\small
\caption{Number of the problematic \ac{sparql} templates from the two
\acp{tgm} for each of the evaluation criteria (\textSMC{DC}
represents the term ``disconnected'').}
\tablabel{errors}
\begin{tabular}{lrrrrrr}
\ac{tgm} name & \ac{tgm} failure & Syntax & Question type
  & \textSMC{DC} target & Wrong range & \textSMC{DC} triple \\
\hline
Rocknrole  & 0 & 0  & 262 & 330 & 28 & 2,898 \\
\ac{lodqa} & 1 & 18 & 446 & 0   & 64 & 0     \\
\end{tabular}
\end{table}
\begin{figure}[p]
\centering\small
\begingroup\input{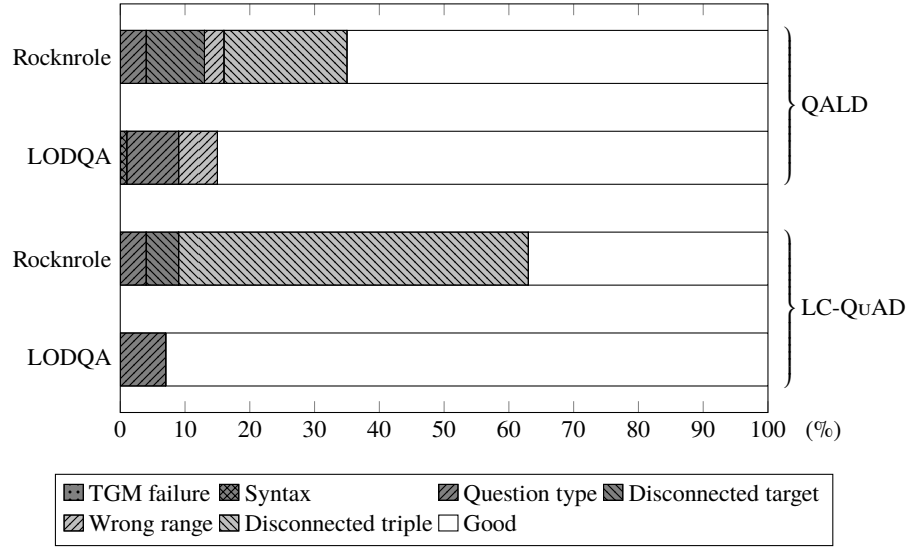}\endgroup
\caption{Ratios of the problematic \ac{sparql} templates from the two
\acp{tgm} for all evaluation criteria. The dark segments~\textlegend{gray}
represent the ratios for the \emph{critical} criteria, and the lighter
segments~\textlegend{gray!50!white} show those for the \emph{notice} criteria.}
\figlabel{errors}
\end{figure}

\section{Discussions}
\seclabel{discussion}

\subsection{Qualitative Evaluation of the \acp{tgm}}

\subsubsection{Rocknrole}

Since Rocknrole is a rule-based \ac{nlq} interpreter, the coverage of the
system is dependent on the rules.  According to our evaluation results, we
determined that (1)~this \ac{tgm} covers both the question types, i.e., yes/no
questions and factoid questions, which are possibly assigned to \ac{sqa}
systems, (2)~the system is also able to add range specification to the template
queries, and (3)~the system, however, often fails to generate good \ac{sparql}
templates. It is worth noting that Rocknrole can perfectly recognize yes/no
questions in our experiments, but it judged 4.7\% of the factoid questions as
yes/no questions throughout all the datasets used herein. In addition, the
\ac{tgm} failed to add appropriate range specification to its outputs for
nearly half of the range-specified factoid questions (45.2\%) in the \ac{qald}
datasets. Note that this level of insight into the performance of the \ac{tgm}
is something that could only be achieved through the subdivided evaluation
proposed in this study.

One possible way to improve Rocknrole to prevent the occurrence of errors,
particularly the \emph{question-type}, \emph{disconnected target}, and
\emph{wrong range} errors, is to add predefined rules to decrease the error
rates. For this purpose, the false cases dumped by our evaluator will be
useful. Furthermore, Rocknrole can only enhance the system to remove
disconnected triples so that such triples will not decrease the values of
recalls.

\subsubsection{\acs*{lodqa}}

As for the \ac{tgm} provided by \ac{lodqa}, the evaluation shows that (1)~it
does not distinguish the yes/no- and factoid-type questions, (2)~it does not
produce range specification to the template queries, and (3)~it, however,
produces good \ac{sparql} templates more stably than Rocknrole. Since the
system does not cover yes/no questions and range-specified factoid questions,
the number of \emph{question-type} and \emph{wrong range} problems presented in
Table~\tabref{errors} and Figure~\figref{errors} are only those problems that
are included in the datasets.

\subsection{Quality of the Datasets}

Figure~\figref{questions} shows that the questions in the \ac{qald} datasets
are richer in diversity than those in \ac{lc-quad}. In fact, \ac{lc-quad} is
not sufficiently diverse to be used solely for our evaluation because it
contains no range-specified factoid questions.

The size of both the datasets seems sufficiently large for our evaluation task
because we were able to find at least one case for each of the criteria.
Nevertheless, the larger the size is, the better the dataset for our
evaluation. In this regard, \ac{lc-quad} has an advantage. The generation
process of \ac{lc-quad} is quite unique: the \ac{nlq} in the dataset is
transformed from \ac{sparql} queries, which is completely opposite to the
generation processes of the other well-known datasets, including
\ac{qald}~\cite{trivedi2017}. As the translations from \ac{sparql} queries to
\acp{nlq} were conducted using specific question templates called ``normalized
natural question templates,'' if the variety of the templates is increased, the
dataset will be more useful for our evaluation.

\subsection{Unsuitable evaluation criteria}
\seclabel{unsuitable}

Several methods have been proposed to determine the similarity degrees or
distances among \ac{sparql} queries~\cite{le2012,dividino2013}, which seems to
be useful for evaluating \ac{sqa} systems. Although this may also be true for
evaluating the whole \ac{okbqa} framework, these methods are not appropriate
for our evaluation because our evaluation currently focuses on only \acp{tgm}.
First, most aforementioned measurements cannot be applied to \ac{sparql}
queries that do not contain \acp{uri}. Second, comparing or measuring the
similarity between the graph patterns of the \ac{sparql} queries without
resource annotations seems meaningless because one semantic structure of a
question can generally be expressed in several forms of \ac{sparql} queries.

Another possible criterion is the expected type of the answer, namely the type
or class of the target variable in the template queries. However, this is also
not appropriate for our evaluation because template queries do not contain any
\ac{uri}. We thus reserve this for the evaluation of other modules, which
follow a \ac{tgm} in a \ac{qa} workflow.

\subsection{Limitations, possible extensions, and future work}

Both \ac{tgm} implementations currently do not generate more than one
\ac{sparql} template for an \ac{nlq}, whereas the specification of the
\ac{okbqa} framework allows \acp{tgm} to generate multiple templates for an
input. For this reason, currently, we simply took the first template from the
result lists, the length of which is always one, returned by \acp{tgm};
however, this behavior is required to be changed for \acp{tgm} that generate
more than one templates for an \ac{nlq}. We think we can always take standard
approaches which are broadly understood. For example, we can count positives/%
negatives and evaluate the performance in terms of precision and recall.

Though we evaluated only two \acp{tgm} for our experiments, our methods and
implementation can be easily applied to other \ac{tgm} implementations by
merely specifying the \ac{url} of the \ac{rest} service. Other datasets from
\ac{qald} and \ac{lc-quad} can also be used for the evaluation. Our evaluation
is not specialized for English questions; thus, if applicable \acp{tgm} and
datasets are provided, these \acp{tgm} can also be tested for other languages.
Furthermore, our evaluation methods can possibly be applied not only to
\ac{tgm} in the \ac{okbqa} framework but also to every \ac{sqa} system that
generates \ac{sparql} queries because the methods are solely based on the
results of a general \ac{sparql} parser. The results in such cases will
demonstrate the performance of the functions corresponding to the \ac{tgm} of
the systems, which will be helpful in improving the systems as well.

Finally, for future research, defining and developing similar subdivided,
systematic, and semantic evaluations of the other modules of the \ac{okbqa}
framework will also be interesting. Evaluation methods optimized for more
complex questions than our current corpus, such as those discussed by Talmor et
al.\@ \cite{talmor2018}, should also be implemented. Simultaneously, these
evaluations will be helpful in improving the ability of the entire \ac{sqa}
system.

\section{Conclusions}

Herein, we proposed a systematic semantic evaluation for \acp{tgm}, which are a
type of subdivided modules of \ac{sqa} systems. Our evaluation results for the
two publicly available \acp{tgm} revealed that in comparison with existing
methods, the new evaluation method can extract and provide much more detailed
information on their performance. Specifically, the limitations and the
problems of the \acp{tgm} were detected; these are hoped to be fixed in the
future. The information from our evaluator will be useful for addressing this
issue.

Improvement of \ac{sqa} systems is important for expanding the use of \ac{ld},
and this paper showed that the presented evaluation method has a good potential
to play an important role for advancing the technology. Therefore, it will be
worth to extend the approach to other modules of \ac{sqa} systems, and even to
other \ac{sqa} frameworks, which is remained as a prospective future work.

\end{document}